\documentclass{article}
\PassOptionsToPackage{numbers, compress}{natbib}
\usepackage[preprint]{neurips_2023}

\usepackage[utf8]{inputenc} % allow utf-8 input
\usepackage[T1]{fontenc}    % use 8-bit T1 fonts
\usepackage{hyperref}       % hyperlinks
\usepackage{url}            % simple URL typesetting
\usepackage{booktabs}       % professional-quality tables
\usepackage{amsfonts}       % blackboard math symbols
\usepackage{nicefrac}       % compact symbols for 1/2, etc.
\usepackage{microtype}      % microtypography
\usepackage{xcolor}         % colors
\usepackage{tabularx}
\usepackage{url}
\usepackage[utf8]{inputenc}
\usepackage{caption}
\usepackage{graphicx}
\usepackage{amsmath}
\usepackage{tabularx}
\usepackage{amsthm}
\usepackage{wrapfig,booktabs}
\usepackage{algorithm}
\usepackage{algorithmic}
\usepackage{graphicx}
\usepackage{amssymb}
\usepackage{booktabs}
\usepackage{multirow}
\usepackage{caption}
\usepackage{subcaption}
\usepackage[capitalize]{cleveref}
\crefname{section}{Sec.}{Secs.}
\Crefname{section}{Section}{Sections}
\Crefname{table}{Table}{Tables}
\crefname{table}{Tab.}{Tabs.}

\title{Overcoming Topology Agnosticism: Enhancing Skeleton-Based Action Recognition through Redefined Skeletal Topology Awareness}

\author{
Yuxuan Zhou$^1$ \hspace{10pt}
Zhi-Qi Cheng$^2$\thanks{Corresponding author} \hspace{10pt}
Jun-Yan He$^3$ \hspace{10pt} \\
\textbf{Bin Luo}$^3$ \hspace{10pt}
\textbf{Yifeng Geng}$^3$ \hspace{10pt}
\textbf{Xuansong Xie}$^3$ \hspace{10pt}
\\
$^1$University of Mannheim,
$^2$Carnegie Mellon University,
$^3$Alibaba Group\\
{\tt \small zhouyuxuanyx@gmail.com,
zhiqic@cs.cmu.edu}\\
{\tt \small \{leyuan.hjy, luwu.lb, cangyu.gyf\}@alibaba-inc.com, xingtong.xxs@taobao.com
}
}

\begin{document}
\maketitle

\begin{abstract}
Graph Convolutional Networks (GCNs) have long defined the state-of-the-art in skeleton-based action recognition, leveraging their ability to unravel the complex dynamics of human joint topology through the graph’s adjacency matrix. However, an inherent flaw has come to light in these cutting-edge models: they tend to \textit{optimize the adjacency matrix jointly with the model weights.}  
This process, while seemingly efficient, causes a gradual decay of bone connectivity data, culminating in a model indifferent to the very topology it sought to map.
As a remedy, we propose a threefold strategy:
(1) We forge an innovative pathway that encodes bone connectivity by harnessing the power of graph distances. This approach preserves the vital topological nuances often lost in conventional GCNs.
(2) We highlight an oft-overlooked feature - the temporal mean of a skeletal sequence, which, despite its modest guise, carries highly action-specific information.      
(3) Our investigation revealed strong variations in joint-to-joint relationships across different actions. This finding exposes the limitations of a single adjacency matrix in capturing the variations of relational configurations emblematic of human movement, which we remedy by proposing an efficient refinement to Graph Convolutions (GC) - the \textit{BlockGC}. 
This evolution slashes parameters by a substantial margin (above $40\%$), while elevating performance beyond original GCNs. Our full model, the \textit{BlockGCN}, establishes new standards in skeleton-based action recognition for small model sizes. Its high accuracy, notably on the large-scale NTU RGB+D 120 dataset, stand as compelling proof of the efficacy of \textit{BlockGCN}. 

\end{abstract} 

\maketitle
\section{Introduction}
\label{sec:intro}
The realm of skeleton-based action recognition has undergone a transformative evolution, born out of the need for computational efficiency, and adaptability to varying environmental conditions, particularly in fields such as medical applications.
Initial approaches leaned heavily on Recurrent Neural Networks (RNNs) and Convolutional Neural Networks (CNNs), employing features or pseudo-images derived from human joints to generate predictions. While performing well in general, these methods are limited in capturing the intrinsic correlations that exist between human joints - a fundamental prerequisite for nuanced human action recognition.

Graph Convolutional Networks (GCNs) \cite{defferrard2016convolutional, kipf2016semi, schlichtkrull2018modeling} 
 facilitate to overcome such issues by representing joints and their physical connections as nodes and edges of a graph respectively, thus making the analysis of joint interactions more dynamic and meaningful. However, this manually defined topology of GCNs overlooks relationships between physically unconnected joints, thus inadvertently limiting the representational power. Furthermore, the predetermined connections can not quite incorporate the hierarchical structure of GCNs, which aimed to capture multi-level semantic information.

Learnable topologies can address this issue and have demonstrated impressive adaptability and flexibility (e.g.~\cite{chen2021channel, cheng2020decoupling}). They are usually initialized as the natural skeleton, following the intuition to provide this topological information to the network, yet have an obvious practical trade-off- the explicit skeletal topology encoding can gradually be eroded during training. In fact, our detailed examination empirically shows that this valuable topology information, initially provided in the learnable adjacency matrix, tends to fade during training, reducing its significance in such fully-connected adjacency models to a mere initialization tool. Consequently, the network's ability to harness relative spatial information between neighboring joints deteriorates. The skeletal topology, along with the essential positional information on body joints, becomes increasingly elusive as GCN training progresses.

Our remedy for this predicament is a novel approach that we term \emph{Topological Invariance Encoding}. In this encoding, the skeletal topology is expressed through relative distances between pairs of joints on the skeletal graph, leading to a more accurate and sustainable representation of the skeletal structure. Complementing this Topological Invariance Encoding, we have developed Statistical Invariance Encoding, which exploits a statistical invariant positional feature - the relative coordinate distances between joints of the average frame - that provides crucial insights of human skeletal structure in addition to graph distances. Our exploration also reveals that joint-to-joint relations are far from static, with considerable variations across different actions.

In response to this finding, we propose a significant refinement to the conventional Graph Convolution (GC) - BlockGC. This novel extension proves to be a tour de force in terms of both efficiency and performance, adept at multi-relational modeling, and \textbf{reducing parameters by almost half (43\%) while boosting performance}. Our key contributions are:
\begin{itemize}
    \item The identification and rectification of the skeletal topology oversight in state-of-the-art GCNs, achieved through our novel Topological Invariance Encoding.
    \item The introduction of Statistical Invariance Encoding, a method that harnesses the temporal average of a pose sequence, providing a robust defense against noise.
    \item The development of BlockGC, an efficient and powerful extension of Graph Convolution (GC), that {decreases parameters by nearly half while boosting performance}, due to its block diagonal weight matrix.
    \item The establishment of new performance benchmarks on the large-scale NTU RGB+D 120 dataset, courtesy of our proposed methods.
\end{itemize}

\begin{figure}[t]
     \centering
    \begin{subfigure}[t]{0.65\textwidth}
         \centering
     \includegraphics[width=\textwidth]{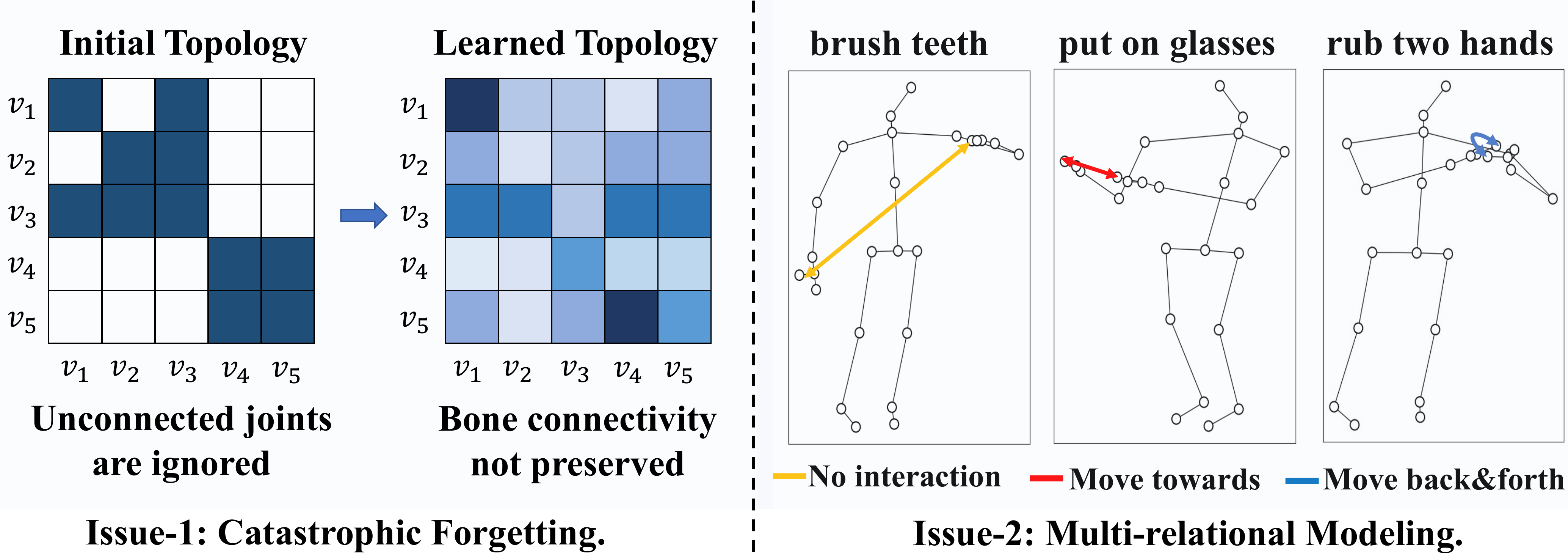}
     \caption{Unresolved issues of  GCN-based approaches. Skeletal information is lost after training (left) and joint relations vary in different actions (right).}
         \label{fig:tease}
     \end{subfigure}\hfill
     \begin{subfigure}[t]{0.33\textwidth}
         \centering
         \includegraphics[width=0.85\textwidth]{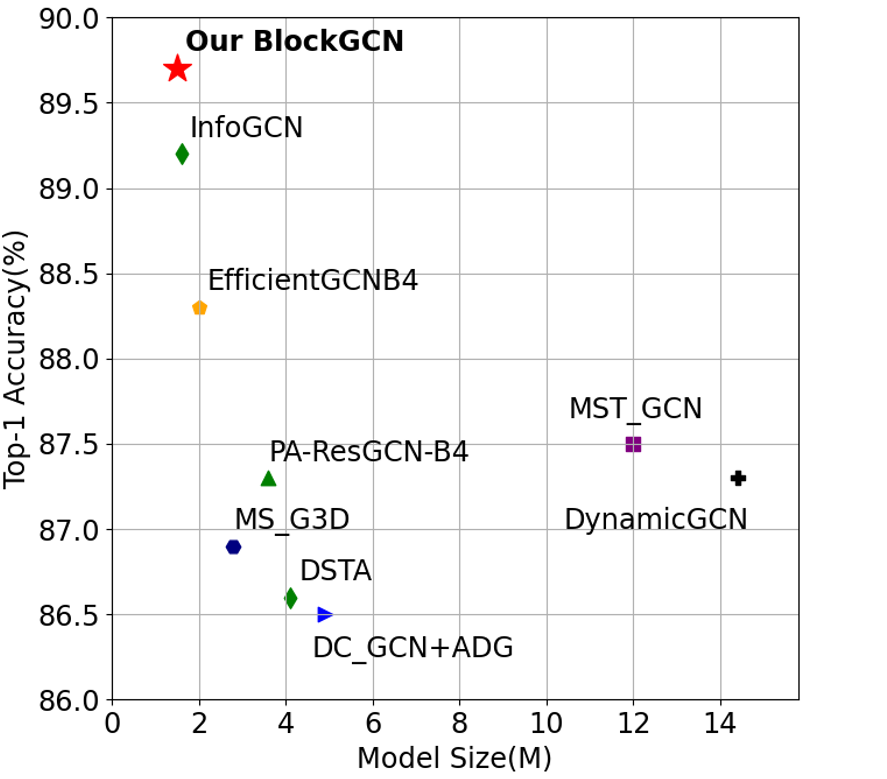}
         \caption{Performance vs. Model Size on NTU RGB+D 120 Cross-Subject.}
         \label{fig:performance}
     \end{subfigure}\hfill
    \caption{We reveal the remaining issues of previous GCNs in \cref{fig:tease} and propose BlockGCN as the remedy, which improves over previous methods w.r.t. both performance and efficiency, see \cref{fig:performance}.}
\end{figure}

\section{Related Work}
\label{sec:related}
\subsection{Traditional Approaches to Skeleton-based Action Recognition}
Early approaches to skeleton-based action recognition relied on Recurrent Neural Networks (RNNs) due to their ability to handle temporal dependencies \cite{du2015hierarchical, song2017end, zhang2017view}. Convolutional Neural Networks (CNNs) were also used, but they were found to be less effective in explicitly capturing spatial interactions among body joints \cite{ke2017a, liu2017enhanced}. As a result, the focus shifted to Graph Convolutional Networks (GCNs), which extended convolution operations to non-Euclidean spaces and enabled the explicit modeling of joint spatial configurations \cite{gilmer2017neural, xu2018how}. \textit{In the following, we primarily focus on these graph-based models as they more comprehensively capture spatial relationships.}

\subsection{Graph Convolutional Networks for Skeleton-based Action Recognition}
Graph Convolutional Networks (GCNs) by Kipf and Welling \cite{kipf2016semi} have had a significant impact on skeleton-based action recognition. \textit{However, GCN-based methods have certain limitations:}

\textbf{1)~Choice of Topology}. The choice of graph topology in GCNs is crucial. Early works, such as Yan et al. \cite{yan2018spatial}, used a fixed topology based on bone connectivity, demonstrating the effectiveness of GCNs in action recognition. However, this rigid topology has inherent limitations. Recent approaches have explored learnable adjacency matrices to capture relationships between physically connected and unconnected joints \cite{chen2021channel, cheng2020decoupling, chi2022infogcn, gao2022global, liu2020disentangling, shi2019two, song2021constructing, xia2021multi, ye2020dynamic}. Our work builds on this idea and addresses Catastrophic Forgetting associated with learnable adjacency matrices, proposing a method to preserve bone connectivity information.

\textbf{2)~Relative Positional Encodings}. Relative positional information has proven important in various domains, including Natural Language Processing \cite{dai2019transformer, he2020deberta, shaw2018self} and Computer Vision \cite{wu2021rethinking, zhou2022sp}. While relative positional encoding has been demonstrated beneficial for  Transformers on graph data \cite{ying2021transformers}, its significance for GCNs, and especially in the field of skeleton-based action recognition, remains unexplored. Our work aims to fill this gap by proposing a novel method for relative positional encoding that preserves spatial and temporal invariances in skeleton data.

\textbf{3)~Multi-Relational Modeling}. Capturing multiple semantic relations with a single adjacency matrix is challenging. Previous studies have proposed strategies to overcome this limitation:
\textit{1)~Ensemble of GCs}: Yan et al. \cite{yan2018spatial} employed three parallel GCs at each layer, with each adjacency matrix derived from the distance to a reference node. However, we observed that each adjacency matrix tends to become fully connected after learning, rendering the handcrafted partitions ineffective. This setup is equivalent to ensembling multiple GCs at each layer, a technique adopted in subsequent work \cite{chen2021channel, cheng2020decoupling, chi2022infogcn, liu2020disentangling, shi2019two, yan2018spatial, zhang2020semantics}.
\textit{2)~Ensemble of Adjacency Matrices}: DecouplingGCN \cite{cheng2020decoupling} uses multiple adjacency matrices for different subsets of feature dimensions, increasing expressiveness at the cost of parameters and computational demand.
\textit{3)~Attention-based Adaptation of Adjacency Matrix}: Recent works \cite{chen2021channel, cheng2020decoupling, chi2022infogcn, gao2022global, shi2019two, ye2020dynamic} incorporate attention mechanisms or similar techniques to create a data-dependent component of the topology, similar to Graph Attention Networks \cite{velivckovic2017graph}. This approach allows for the dynamic adjustment of joint connections based on relevance but is computationally heavy and requires extensive data for optimal performance. In contrast to the above mentioned approaches, our proposed BlockGC enables the full power of multi-relational modeling by assigning a unique subset of weights to each feature group, at the same time being the most efficient by defining a sparse projection weight matrix.

\section{Method}
\label{sec:method}
In this work, we initially juxtapose Graph Convolutional Networks (GCNs) that utilize learnable adjacency matrices with Fully Connected Networks (FCNs). Through a combination of theoretical and experimental analyses, we identify two primary challenges: \textit{1)~catastrophic forgetting of skeletal topology} and \textit{2)~insufficient capacity to learn joint co-occurrences} (Sec.~\ref{sec:revisit}). To combat these limitations, we introduce a series of enhancements: \textit{1)~topological and statistical invariance encoding} aimed at retaining key skeleton properties (Sec.~\ref{sec:encoding}), and \textit{2)~an enhanced graph convolution}, termed BlockGC, designed to capture the implicit relations within joints (Sec.~\ref{sec:blockgc}).
The above innovations lead to the core building block of our Model, as shown in (see~\cref{fig:main} (bottom)). 

\subsection{Reassessing the Limitations of GCNs}
\label{sec:revisit}
Within the realm of skeleton-based action recognition, the human body's topology is inherently defined as a graph $\mathcal{G} = (\mathcal{V}, \mathcal{E})$, where the vertices $\mathcal{V}$ represent the body's joints, and the edges $\mathcal{E}$ illustrate the connections between joints through bones. As a result, nearly all cutting-edge methods\cite{chen2021channel, cheng2020decoupling, liu2020disentangling, shi2019two, song2021constructing, xia2021multi, ye2020dynamic} consistently adopt the graph convolution,
\begin{equation}
\label{eq:1}
H^{(l)} = \sigma(A^{(l)}H^{(l-1)}W^{(l)})\, ,
\end{equation}
where $A^{(l)} \in \mathbb{R}^{V \times V}$ is the adjacency matrix employed for spatial aggregation, $H^{(l)} \in \mathbb{R}^{ V\times T \times D}$ symbolizes the hidden representation, and $W^{(l)} \in \mathbb{R}^{D \times D}$ is the weight matrix utilized for feature projection. Here, $V$, $T$, and $D$ denote the number of joints, frames, and hidden features, respectively. $\sigma$ is the non-linear ReLU activation function, and the superscript $l$ indicates the layer number. Despite GCNs seeming adept at learning human skeleton characteristics effectively, \textit{our experimental validation shows that this is not entirely the case.} 
To sum up, there are two main issues in existing GCNs, which will be systematically analyzed below.

\noindent \textbf{Problem-1: Catastrophic Forgetting of skeletal topology}.~Prior research can generally be categorized into two groups: one\cite{yan2018spatial} where the adjacency matrix is fixed to portray the skeleton topology, and the other\cite{chen2021channel,cheng2020decoupling, chi2022infogcn, shi2019two} where the adjacency matrix is optimized during training via gradient backpropagation\footnote{For details, please refer to related work.}. Despite these advancements, GCNs (Eqn.~\ref{eq:1}) have been observed to struggle with accurate recognition of complex actions\cite{cheng2020decoupling}. We hypothesize that this performance bottleneck is related to the adjacency matrix $A$, as it "catastrophically forgets" the skeleton topology during training. Our goal is to validate this hypothesis through both theoretical and experimental approaches.

Theoretically, GCNs can be interpreted as a fully connected layer with a weight matrix $W_{spatial} \in \mathbb{R}^{V \times V}$. In this light, GCNs resemble ResMLP~\cite{touvron2021resmlp} and MLP-Mixer~\cite{tolstikhin2021mlp}, which are typically employed for image classification. However, both ResMLP and MLP-Mixer have been shown to suffer from catastrophic forgetting~\cite{parisi2019continual} during training, resulting in the inability to preserve the original topological representation in the adjacency matrix $A$.

From an experimental perspective, we have rigorously confirmed the catastrophic forgetting of skeleton topology (\cref{tab:pre}). Our results demonstrate that GCNs' performance remains similar irrespective of the initialization states, suggesting that existing GCNs entirely fail to maintain the topological skeleton in the adjacency matrix $A$. Additionally, our supplementary visualization and statistical analysis also corroborate this conclusion.

\noindent \textbf{Problem-2: Insufficient capacity to learn joint co-occurrences}.~The interactions between joints are action-dependent. For instance, during running, the movement of hands and feet primarily serves to maintain balance, whereas when removing shoes, hands and feet interact more directly and play a dominant role. Therefore, it is clear that a single adjacency matrix $A$ in a classic GCN (Eqn.~\ref{eq:1}) cannot capture more than one type of interaction. 

\begin{figure}[ht]
\begin{subfigure}{0.45\textwidth}
\centering
\includegraphics[width=\textwidth]{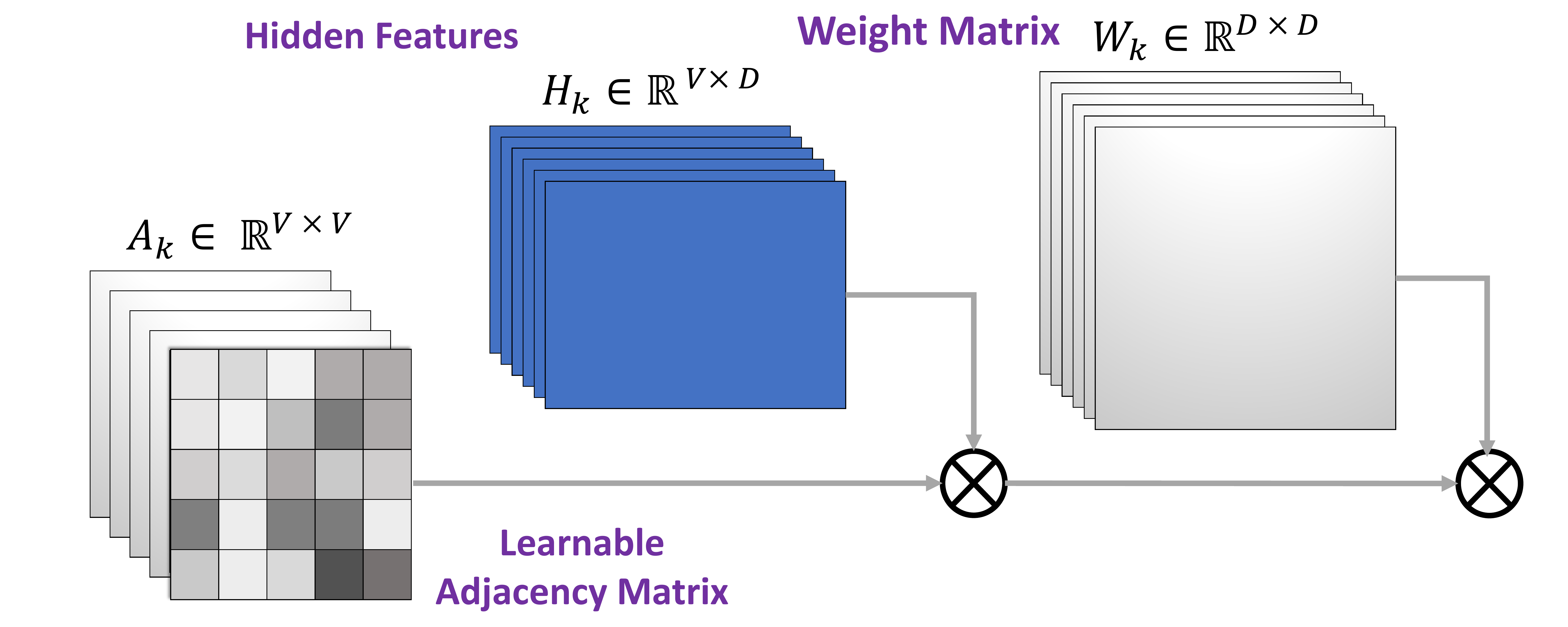}
  
    \caption{Ensemble of GCs.}
  \label{fig:compare1}
   \end{subfigure}
   \begin{subfigure}{0.45\textwidth}
\centering
\includegraphics[width=\textwidth]{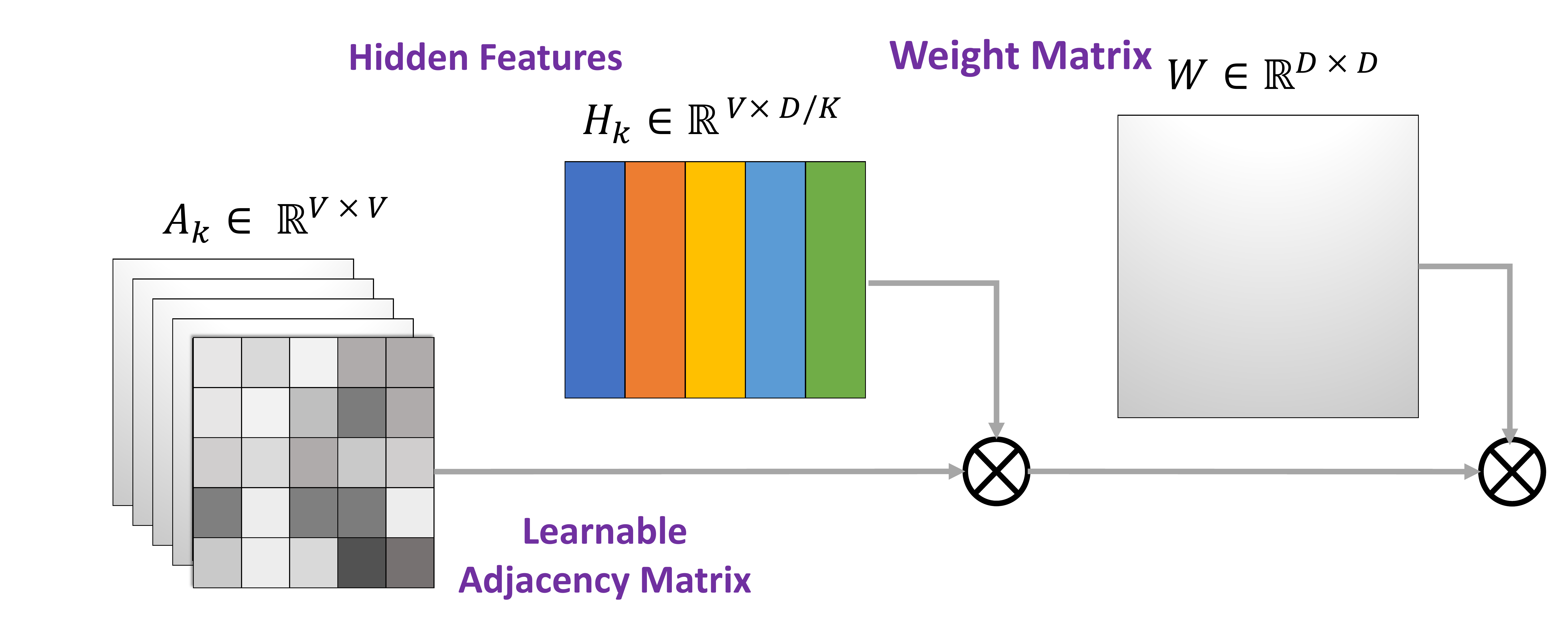}
    \caption{Ensemble of Adjacency Matrices.}
  \label{fig:compare2}
   \end{subfigure}
\begin{subfigure}{\textwidth}
    \centering
\includegraphics[width=0.82\textwidth]{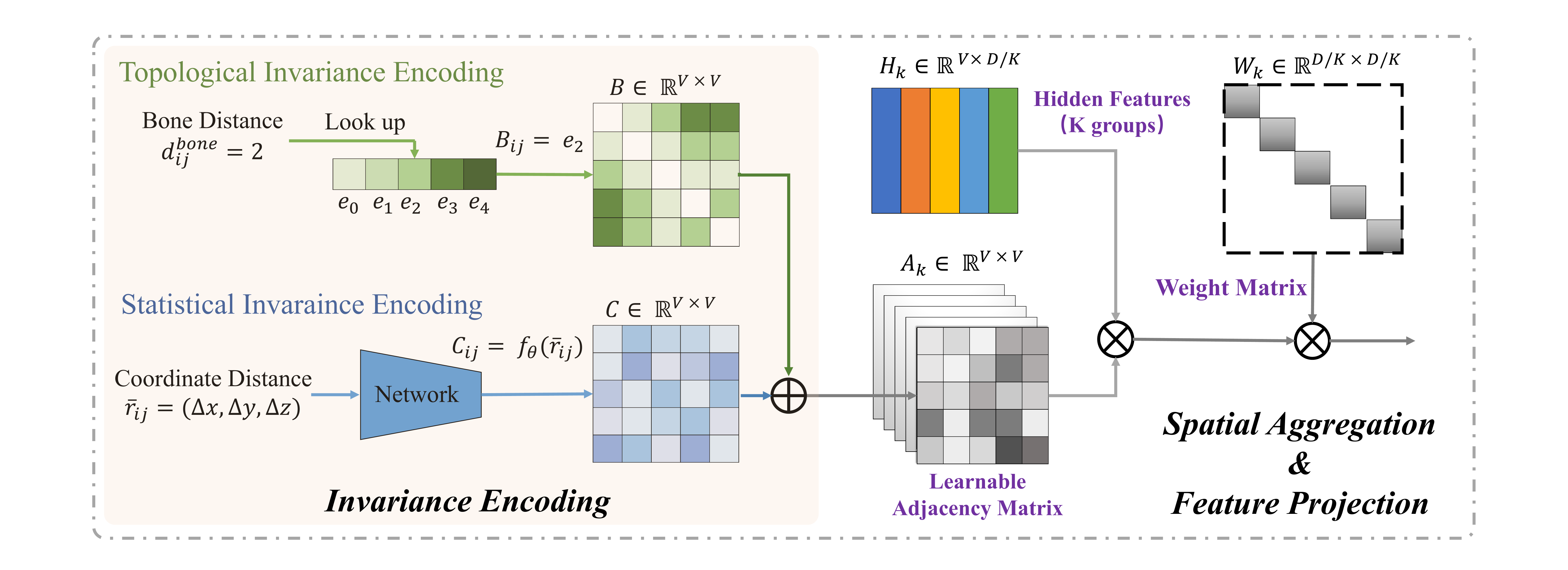}
     \caption{Our BlockGC with invariance encoding. 
     }
    % \caption{}
  \label{fig:blockgc}
\end{subfigure}
   \caption{Illustration of existing approaches for multi-relational modeling (top) and our proposed BlockGC with Invariance Encodings (bottom). Invariance Encodings preserve the information of skeletal structure, while BlockGC enables multi-relational modeling, at the same time slashing the redundant weights for feature projection, thanks to its block diagonal projection matrix.\label{fig:main}}
   \label{fig:compare}
\end{figure}

To overcome this issue, previous work has proposed the use of an ensemble of GCs, i.e., an ensemble of adjacency matrices, and the adaptation of the adjacency matrix (see~\cref{fig:main} (top)). For layer-wise ensembles of GCs, both parameters and computation increase linearly with the number of ensembles, causing the model to become excessively large with many ensembles and to suffer from over-fitting. As a result, the number of ensembles is typically limited to three. 

For the ensemble of adjacency matrices \cite{cheng2020decoupling} and attention-based adaptation \cite{chen2021channel,chi2022infogcn}, a single weight matrix is applied across the entire feature dimension, which constrains the modeling capacity. Furthermore, our experimental results demonstrate that a significant portion of the weight matrix is redundant (see~\cref{tab:multi-rel}).

\subsection{Topological and Statistical Invariance Encoding}
\label{sec:encoding}
GCNs with trainable adjacency matrices $A$ become insensitive to the underlying skeletal topology, i.e., the bone connections, post-training. Nevertheless, access to bone connections is beneficial since they convey substantial information about the action being performed, such as how the bone connections physically limit joint movements. To preserve this information, we introduce a method termed Topological Invariance Encoding. Moreover, we consider another approximately invariant feature, namely the mean frame of a pose sequence, which provides a significant clue (see Fig.~\ref{fig:temporal}). To incorporate this feature into our model, we propose a technique called Statistical Invariance Encoding.

\subsubsection{Encoding the topological invariance}
Bones connect the human body's joints, which physically restrict each joint's movement during an action. It is critical to integrate this bone connectivity to recognize the action. We suggest a method called Topological Invariance Encoding to include the skeletal topology. This method encodes the relative distance between two joints on the skeletal graph $\mathcal{G}$, using different distance measures such as shortest path distance or distance in a level structure \cite{diaz2002survey}. Due to its simplicity, we adopt the shortest path distance for our final model.
\begin{equation}
         B_{ij} = e_{d_{i,j}^{\mathit{bone}}}
        \quad \mathrm{with} \quad
 d_{i, j}^{\mathit{bone}}  = \min\limits_{P \in \mathit{Paths}(\mathcal{G})} \{\vert P\vert, P_1 = v_i, P_{|P|} = v_j\}   ,
\end{equation}
where a weight parameter $B_{ij}^{(l)}$ is assigned from a parameter table $E=\{e_{\text{index}}\}$ to each joint pair according to their shortest path length $d_{i,j}^{\mathit{bone}}$ through bone connections.

\subsubsection{Encoding the statistical invariance}
\begin{figure}[ht]
     \centering
         \centering
         \includegraphics[width=0.9\textwidth]{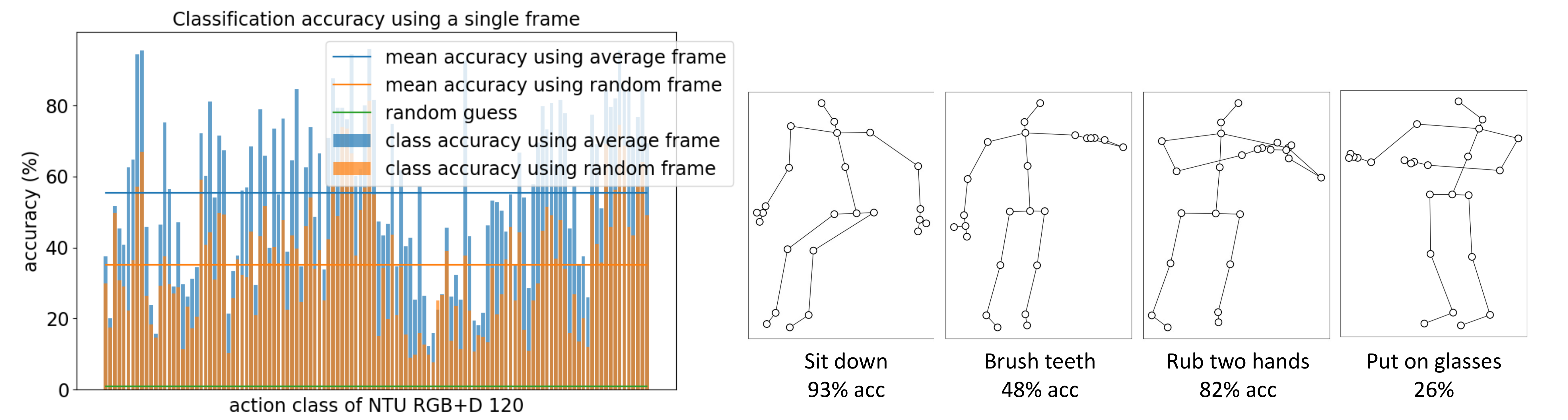}
        \caption{Visualization of the classification accuracy on a single frame of NTU RGB+D 120 Dataset samples. We train a model w/o temporal module only on a single frame, i.e., the temporally averaged frame, compared to one randomly sampled frame. The former has a much higher accuracy than the latter ($54\%$ vs. $35.5\%$), and they are both higher than random guesses ($\frac{1}{120}$). }
        \label{fig:temporal}
\end{figure}

In addition to the skeletal topology, temporally invariant features can offer valuable insights about action, as they are robust to noise. For instance, cyclic joint movements are frequently involved in an action, and the expectation of these patterns over one period represents the temporally invariant feature. Moreover, human joints are physically constrained, and their movements often follow a back-and-forth pattern, whether cyclic or not. Therefore, we can compute the mean of a pose sequence as an approximation of the temporally invariant feature. As shown in \cref{fig:temporal}, such a feature conveys surprisingly rich information about the action class.

To leverage this information, we propose to first calculate the temporal mean of the relative coordinates between each joint pair to obtain $\bar{r}_{ij} \in \mathbb{R}^ 3$. We then encode this mean value to its corresponding weight $C_{ij}$ for each joint pair at each layer through a mapping $f_\theta: \mathbb{R}^3\to\mathbb{R}^D$.% parameterized by a two-layer MLP.
\begin{align}
     \quad  C_{ij} = f_\theta
    (\bar{r}_{ij})  \quad \mathrm{with} \quad \bar{r}_{ij} = \frac{\sum_{t=0}^T (r_{ij}^t)}{T},
\end{align}
where $r_{ij}^t \in \mathbb{R}^{3}$ denotes the coordinates between the $i^{th}$ and $j^{th}$ joints at the $t^{th}$ time frame, and $C \in \mathbb{R}^{V \times V \times D}$ represents the encoded weights for each joint pair. $f_\theta$ is parameterized by a two-layer MLP. Note that the last dimension $D$ is designed to assign a unique encoding to each feature dimension, which has been proven to be more powerful than a shared encoding (as shown in Tab.~\ref{tab:statistical encoding}).

Finally, we sum the learnable adjacency matrix $A \in \mathbb{R}^{V \times V}$ and our Invariance Encoding, which includes both topological and statistical components, to obtain the final matrix for spatial aggregation: 
\begin{equation}
    H^{(l)} = \sigma((A^{(l)} + B^{(l)} + C^{(l)})H^{(l-1)}W^{(l)}).
\end{equation}

\subsection{Learning Multi-Relational Semantics}
\label{sec:blockgc}
Joint co-occurrences inherently involve multiple relations, as discussed in \cref{sec:revisit}, which necessitate the modeling of various semantics. A single adjacency matrix is insufficient to handle such complexity. Previous approaches, detailed in \cref{sec:related}, have limitations in computational efficiency or theoretical constraints, preventing the full potential of GCNs from being realized. To overcome this, we propose a method called BlockGC, allowing fully decoupled modeling of different high-level semantics. Our proposed BlockGC not only reduces computation and parameters but also proves to be more effective than previous methods.

As illustrated in \cref{fig:main} (bottom right), the feature dimension is divided into $K$ groups. Spatial aggregation and feature projection are then applied in parallel on each $k^{th}$ group.

\begin{equation}
H^{(l)}=
\sigma (\left[ 
\begin{array}{c}
 (A_1+ B_1 + C_1)H_1^{(l-1)} \\
 \dots \\
                (A_k+ B_k + C_k)H_k^{(l-1)} \\
                \dots
                \end{array}
                \right]
                \left[
\begin{array}{cccc}
W_1^{(l)}& \\
&\dots &  \\ 
 & & W_k^{(l)}& \\ 
 && & \dots
\end{array}\right])
\end{equation}
where $H_k \in \mathbb{R}^{V \times T \times D/K}$ and $W_k \in \mathbb{R}^{D/K \times D/K}$. $\{W_k, k=1,...,K\}$ are arranged as a block diagonal matrix, which not only leads to parameter reduction but also makes the projected feature groups independent from each other. This is a desired property, as each group is intended to model a kind of semantics that are also independent of each other. 
Thanks to the decoupled feature projection, our method enables GCN the full power for multi-relational modeling. Compared to DecouplingGCN \cite{cheng2020decoupling} and attention-based adaptation of adjacency matrix, our BlockGC not only significantly reduces parameters and computation (BlockGC $\mathcal{O}(\frac{VD^2}{K})$, GC $\mathcal{O}(VD^2)$, Decoupling GC $\mathcal{O}(VD^2)$), but also leads to improved performance.

\subsection{Model Network Architecture}

\begin{figure}
    \centering
     \includegraphics[width=0.95\textwidth]{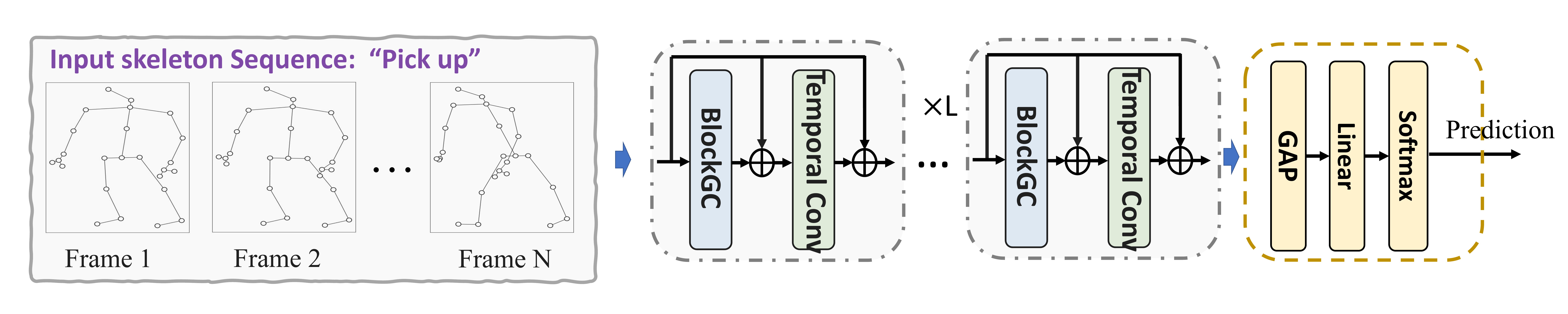}
  \label{fig:arch}
\caption{Model architecture of our BlockGCN. BlockGC captures the joint co-occurrences in the spatial dimension, whereas Temporal Convolution learns the temporal correlations.}
\end{figure}

We built our final model, named BlockGCN, based on the above-described Invariance Encodings and BlockGC, as illustrated in \cref{fig:arch}. 
To model the temporal correlation of the skeleton sequences, we employ the Multi-Scale Temporal Convolution (MS-TC) module \cite{chen2021channel,chi2022infogcn,liu2020disentangling}. It consists of three convolution branches with a 1 × 1 convolution for dimension reduction and different combinations of kernel sizes and dilations. The outputs of convolution branches are concatenated as the final output.

We build our final model by stacking our BlockGC and MS-TC modules alternately 10 times as follows (the Invariance Encodings are omitted for simplification):
\begin{align}
    & H^{(l)} = \text{BlockGC}(H^{(l-1)}) + H^{(l-1)}, \\
    & H^{(l)} = \text{MS-TC}(H^{(l)}) + H^{(l-1)}, \\
    & H^{(l)} = \text{ReLU}(H^{(l)}).
\end{align}
The final output of our model is produced by applying a global pooling operation over both the joint and temporal dimensions, followed by a softmax operation over the class labels. This final model, named BlockGCN, is designed to efficiently and effectively model the multi-relational semantics inherent in human action recognition tasks.

\section{Experiments}
\vspace{-2mm}
In this section, we undertake a comprehensive evaluation of our proposed BlockGCN on standard benchmarks for skeleton-based action recognition. Our empirical results showcase that our model either matches or exceeds the performance of existing state-of-the-art methods such as those presented in \cite{chen2021channel,chi2022infogcn}. Furthermore, we present an intricate analysis exploring the significance of topological information within GCN-based models for action recognition. We also carry out an ablation study to assess the efficacy of our novel Topological and Statistical Invariance Encodings and BlockGC. Remarkably, we employ the standard cross-entropy loss in all our experiments to ensure an impartial assessment of our architecture and to uphold direct comparability with prior works. We gauge the performance of our BlockGCN on three widely-used benchmark datasets for skeleton-based human action recognition: NTU RGB+D~\cite{shahroudy2016ntu}, NTU RGB+D 120~\cite{liu2019ntu}, and Northwestern-UCLA~\cite{wang2014cross}.

\subsection{Implementation Details}
\vspace{-2mm}
We conducted all experiments on a Tesla V100 GPU using the PyTorch deep learning framework \cite{paszke2019pytorch}. To ensure stability during the early training phase, we utilized a warmup technique \cite{he2016deep} for the initial 5 epochs out of a total of 140 training epochs. The model was optimized via Stochastic Gradient Descent (SGD) with Nesterov momentum set at 0.9 and a weight decay of 0.0004 for NTU RGB+D and NTU RGB+D 120, and 0.0002 for Northwestern-UCLA. Our experiments employed cross-entropy loss and initiated the learning rate at 0.1, reducing it by a factor of 10 at epochs 110 and 120, in accordance with the strategy used in \cite{chi2022infogcn}. For NTU RGB+D and NTU RGB+D 120, we opted for a batch size of 64, resized each sample to 64 frames, and adhered to the data pre-processing steps outlined in \cite{zhang2020semantics}. For Northwestern-UCLA, we selected a batch size of 16 and followed the data pre-processing strategies from \cite{chen2021channel, cheng2020skeleton}. Our implementation builds upon the official code \cite{chen2021channel, zhang2020semantics}.

\subsection{Comparison with State-of-the-art}
\label{sec:sota}
\vspace{-2mm}
To establish a fair comparison, we employed the commonly accepted 4-stream fusion approach in our experiments. In particular, we input four different modalities: \textit{joint}, \textit{bone}, \textit{joint motion}, and \textit{bone motion}. The joint and bone modalities denote the original skeleton coordinates and their derivatives with respect to bone connectivity, respectively. The joint and bone motion modalities compute the temporal differential of the joint and bone modalities. Subsequently, we amalgamate the predicted scores of each stream to produce the final fused results. For a fair evaluation, we only consider the results of InfoGCN \cite{chi2022infogcn} that utilize 4 modalities.

We juxtapose our BlockGCN with state-of-the-art methods on NTU RGB+D, NTU RGB+D 120, and Northwestern-UCLA in \cref{tab:ntu}. It is noteworthy that the recently published works \cite{chi2022infogcn,duan2022revisiting,gao2022global} are not directly comparable to our method. \cite{duan2022revisiting} achieves improved results by incorporating additional RGB input, but this necessitates significant computational overhead. InfoGCN \cite{chi2022infogcn} employs an extra loss, which is orthogonal to the design of the architecture. For a balanced comparison, we limit our comparison to their results using the ensemble of 4 modalities. Furthermore, GL-CVFD \cite{gao2022global} has four times larger than ours (6.5M vs. 1.6M parameters), and they rely on a two-stage training strategy.

Our BlockGCN shines in performance on the challenging NTU-RGB+D 120 Cross-Subject benchmark, achieving an accuracy of $89.7\%$ as presented in \cref{tab:ntu}. This result denotes an improvement of $0.5\%$ over the state-of-the-art \cite{chi2022infogcn}, further testifying to the efficacy of our approach.

\begin{table*}[t]
  \centering
  \scriptsize
    \caption{Comparison of BlockGCN and other state-of-the-art methods on NTU RGB+D, NTU RGB+D 120, and Northwestern-UCLA datasets using standard 4 modalities. InfoGCN \cite{chi2022infogcn} reported their results using an ensemble of 6 modalities, but we were unable to reproduce these results using only 4 modalities as per the supplementary materials. This discrepancy has also been publicly acknowledged by Huang et al. \cite{huang2023graph} (see their Tab. 2). Therefore, for a fair comparison, we present our reproduced results for InfoGCN. Refer to \cref{sec:sota} for more details.}
  \begin{tabular}{@{}l|lcccccc@{}}
    \toprule
   \multirow{2}*{Type} & \multirow{2}*{Methods} & \multirow{2}*{Parameters(M)} &\multicolumn{2}{c}{NTU RGB+D 60} &  \multicolumn{2}{c}{NTU RGB+D 120} & \multirow{2}*{NW-UCLA} \\
         &   &  & X-Sub(\%) & X-View(\%)    &  X-Sub(\%)  & X-Set(\%) &  \\
    \midrule
         \multirow{2}*{Transformer} 
    & ST-TR \cite{plizzari2021spatial} & 12.1M & 89.9 & 96.1    & 82.7 & 84.7 & - \\
   & DSTA \cite{shi2020decoupled} & 4.1M & 91.5 & 96.4 & 86.6 & 89.0 & - \\
   \hline
        \multirow{4}*{\shortstack{Hybrid Model \\ (
        GCN + Att})} 
         & SGN \cite{zhang2020semantics} & 0.7M & 89.0 & 94.5 & 79.2 & 81.5 & -\\
       & PA-ResGCN-B19 \cite{song2020stronger}  &3.6M & 90.9  & 96.0 & 87.3  & 88.3  & - \\
       & Dynamic GCN \cite{ye2020dynamic}  & 14.4M &  91.5 & 96.0 & 87.3 & 88.6 & - \\
       & EfficientGCN-B4 \cite{song2021constructing} & 2.0M & 91.7 & 95.7 & 88.3 & 89.1  & - \\
          & InfoGCN* \cite{chi2022infogcn} & 1.6M & 92.3* & 96.5* & 89.2* & 90.6* & 96.5*\\
       \hline
           \multirow{4}*{GCN}
      & DC-GCN+ADG \cite{cheng2020decoupling} & 4.9M & 90.8 & 96.6 & 86.5 & 88.1 & 95.3 \\ 
       & MS-G3D \cite{liu2020disentangling}  & 2.8M & 91.5 & 96.2  & 86.9 & 88.4 & - \\
      
       & MST-GCN  \cite{chen2021multi} & 12.0M& 91.5 & \textbf{96.6} & 87.5 & 88.8 & -\\
               & BlockGCN (ours) & 1.5M & \textbf{92.8}  & 96.4 & \textbf{89.7} & \textbf{90.9} & \textbf{96.6} \\
    \bottomrule
  \end{tabular}
  \label{tab:ntu}
\end{table*}

\subsection{Ablation Analysis}
In this subsection, we delve into an experimental evaluation of the effectiveness of each component of our proposed method. All ablation studies are carried out on the X-sub benchmark of NTU RGB+D 120, utilizing a single joint modality. We initiate the study by examining the impact of different initializations for the adjacency matrix.

% \paragraph{
\begin{wraptable}{r}{0.45\textwidth}%[ht]
\scriptsize
  \centering
    \caption{Ablation on the adjacency matrix initialization on NTU RGB+D 120, X-sub.}
  \begin{tabular}{@{}lc@{}}
    \toprule
    Initialization of Adjacency Matrix & Acc(\%) \\
    \midrule
    Physical Connections \cite{chen2021channel} & 83.9 \\
    Identity Matrix &  \textbf{84.0} \\
    Ones &  83.8 \\
    Kaiming Uniform & 83.8 \\
    \bottomrule
  \end{tabular}
  \label{tab:pre}
% \begin{wraptable}{r}
% {0.45\textwidth}%[ht]
\vspace{0.6 cm}
\centering
   \caption{Ablation on our proposed BlockGC and invariance Encodings. \label{tab:ablation}}
   \scriptsize
  \begin{tabular}
{@{}c@{\hspace{0.2cm}}c@{\hspace{0.1cm}}@{\hspace{0.2cm}}c@{\hspace{0.2cm}}c@{\hspace{0.2cm}}c@{}}
    \toprule
   \multirow{2}*{BlockGC} & \multicolumn{2}{c}{Encoding} & \multirow{2}*{Params} & \multirow{2}*{Acc(\%)} \\
      & statistical & topological & & \\
    \midrule
   - &- & - & 2.1M& 85.2 \\
      \checkmark & - & - & 1.2M & 85.5 \\
        \checkmark & - & \checkmark  &  1.2M & 85.7 \\
      \checkmark& \checkmark & - & 1.5M &  85.9 \\
     \checkmark& \checkmark & \checkmark & 1.5M (-0.6M)  & \textbf{86.0} (+0.8) \\ 
    \bottomrule
  \end{tabular}
\end{wraptable}
% \end{wraptable}
\noindent \textbf{Implications of Adjacency Matrix Initialization}.
\label{sec:pre}
We scrutinize various strategies for initializing the adjacency matrix, ranging from special initialization leveraging physical connections as in \cite{chen2021channel}, to more topology agnostic approaches. For this experiment, we engage a robust baseline model proposed in \cite{chen2021channel}, which demonstrated exceptional performance on the X-sub benchmark of NTU RGB+D 120, employing basic GCN layers with a learnable topology. The experimental setup, barring the initialization, is kept precisely as outlined in \cite{chen2021channel}. Our results suggest that simply initializing the adjacency matrix based on physical connections does not suffice to exploit the skeletal topology effectively, thereby inspiring our proposed Invariance Encodings to preserve such information.

% \paragraph{
\noindent \textbf{Effectiveness of Individual Components}.
We enhance our baseline by either incorporating the invariance encodings or supplanting the vanilla GC with our BlockGC layers. Our BlockGC substantially reduces the parameters by $0.9M$, while simultaneously improving over the vanilla GC. The introduction of statistical topology marginally increases the parameter count but significantly bolsters performance by $0.4\%$. By integrating our BlockGC with both invariance encodings, we outperform the baseline model by $0.8\%$, while concurrently reducing the parameters by approximately $29\%$ as listed in \cref{tab:ablation}.

% \paragraph{
\noindent \textbf{Shared vs. Feature-wise Encodings}.
In comparison to a shared encoding for all feature dimensions, feature-wise encoding provides a larger capacity at the expense of an increase in parameters.  \begin{wraptable}{p}{0.45\textwidth}
 % \vspace{0.5cm}
  \scriptsize
     \centering
    \caption{Feature-wise vs.~shared Encoding. \label{tab:statistical encoding}}
  \begin{tabular}{@{}ccccc@{}}
    \toprule
      \multirow{2}*{Invariance} & \multicolumn{2}{c}{Encoding Dimension} & \multirow{2}*{Acc(\%)} \\
      & shared & feature-wise &  \\
    \midrule
     \multirow{2}*{Statistical}& \checkmark & - & 85.7  \\
      &  - & \checkmark & 86.0 \\
       \hline
       \multirow{2}*{Topological} & \checkmark & -&   86.0 \\
     & -  & \checkmark  & 85.8 \\
    \bottomrule
  \end{tabular}
  % \vspace{-0.8cm}
\end{wraptable}For our topological invariance encoding, given the simplicity of the graph distance (discrete and one-dimensional), a shared encoding is adequate. Consequently, we simply employ a shared topological invariance encoding. In contrast, the Euclidean distance is continuous and spans three dimensions, necessitating a larger capacity to retain such information. As demonstrated in \cref{tab:statistical encoding}, the effectiveness of shared encoding is restricted and becomes imperceptible after rounding.

\begin{wraptable}{r}{0.45\textwidth}%table}[ht]
   \centering
   \scriptsize
     \caption{Comparing different graph distances for topological invariance encoding.}
  \begin{tabular}{@{}cccc@{}}
    \toprule
      \multicolumn{2}{c}{Graph Distance}  & Acc(\%) \\
     shortest path distance & level difference \\
    \midrule
      - & - & 85.7  \\
      \checkmark &  - &  86.0 \\
      - & \checkmark & 86.0 \\ 
    \bottomrule
  \end{tabular}
  \label{tab: topological encoding}
\end{wraptable}

% \paragraph{
\noindent \textbf{Selection of Graph Distance for Topological Invariance
}.~As discussed in \cref{sec:encoding}, we leverage the relative distances between joint pairs on the graph to symbolize graph topology. Theoretically, any proper graph distance could serve this purpose. In our work, we investigate two common types of graph distances for our topological invariance, namely, the shortest path distance and the distance in the level structure \cite{diaz2002survey}. We compare these two distances in \cref{tab: topological encoding}. Interestingly, both distances lead to an equivalent improvement, suggesting that they fundamentally convey the same information, i.e., bone connectivity. To streamline our approach, we default to employing the shortest path distance in our experiments.

\begin{wraptable}{r}{0.45\textwidth}%table}[ht]
\scriptsize
  \centering
  \caption{BlockGC vs.~DecouplingGC~\cite{cheng2020decoupling}. }
  \begin{tabular}{@{}lcccccc@{}}
    \toprule
    Layer & Groups &  Parameters  & Acc(\%) \\
    \midrule
      GC & 1 &  2.1M &     85.2  \\     
  \hline
             & 4 & 2.1M  & 85.5 \\
        \multirow{2}*{DecouplingGC}& 8 & 2.2M  & 85.6  \\ 
      & 16  & 2.3M  & 85.4 \\
      \hline
      \multirow{3}*{BlockGC (ours)} & 4 &  1.2M (-0.9M) & \textbf{85.8}(+0.6)   \\
      & 8   & 1.2M  & 85.5 \ \\
      & 16 & 1.2M  &  \underline{85.7} \\ 
    \bottomrule
  \end{tabular}
  \label{tab:multi-rel}
\end{wraptable}

\noindent \textbf{Contrasting BlockGC with DecouplingGC}.
We pit our BlockGC against DecouplingGC \cite{cheng2020decoupling} in \cref{tab:multi-rel}, using the X-sub benchmark of NTU RGB+D 120. It is important to note that the count of spatial weight parameters inversely correlates with the number of groups, while the number of adjacency matrices increases concurrently. As a result, our BlockGCs with varying groups possess a similar number of parameters. BlockGC significantly trims down the parameters compared to vanilla GC by almost half, yet it still attains a substantial average improvement against the baseline (approximately $0.5\%$). This result is noteworthy as it not only highlights the redundancy in the extensive parameters in the weight matrix for feature projection, but also corroborates our analysis in \cref{sec:blockgc} that the decoupling of features across different groups is a beneficial attribute.

\section{Discussion \& Conclusion}
\noindent\textbf{Broader Impact}.~Skeleton-based action recognition is computationally more efficient compared to video-based action recognition, and therefore finds its application in a broad range of real-world scenarios with limited resources. Additionally, skeleton data erases the identities of human subjects, such that skeleton-based action recognition has a special advantage regarding privacy protection, e.g., for monitoring activities for medical purposes and violent intent detection.

\noindent \textbf{Limitations}.~Our work focuses on the GCNs in Skeleton-based Action Recognition. However, the observation and conclusions are applicable to GCN-based methods on general graph data.

\noindent \textbf{Conclusion}.~We uncover two issues of GCN, namely Catastrophic Forgetting of the skeletal topology and insufficient capacity for modeling multi-relational joint co-occurrences, and propose Invariance Encodings as well as a novel extension of the vanilla GCN to successfully address these issues. Our proposed contributions allow us to significantly reduce the number of model parameters and the training time. The effectiveness of the resulting model is validated by the improved performance on three commonly used benchmarks.
\clearpage

\bibliographystyle{abbrvnat}
\bibliography{acmart}

\newpage
\appendix

\section*{Supplementary Materials}

In supplementary materials, we deliver an in-depth analysis of our BlockGCN's performance, detailing the results obtained from training with each modality in \cref{tab:b}. The \textbf{remarkable improvement} of our method against previous approaches is especially obvious when comparing the results using the \textbf{single modality} in \cref{tab:joint}.
Furthermore, we display the efficacy of our Topological Invariance Encoding normalization strategy in mitigating overfitting in \cref{tab:norm}, thereby further elucidating the design choices underpinning our Topological Invariance Encoding. To assert the statistical significance of our experiments, we report error bars in \cref{tab:rand}.

In addition to quantitative results, we provide a qualitative perspective by displaying the variations in learned adjacency matrices compared to their initial weights based on bone connections. Furthermore, we illustrate the learned weights of our proposed Topological Invariance Encoding, demonstrating the diverse semantic interpretations learned across different GCN layers.

\section{More experiment results}
\label{sec:b}
\subsection{Accuracy using single modalities}
The small performance gaps are not large for all recent approaches mainly because the reported results are an ensemble of 4 modalities, but the real improvement of our method is obvious on the single joint modality (see \cref{tab:joint}).

\begin{table}[ht]
% \scriptsize
  \centering
    \caption{Performance of SOTA methods using joint modality only. * denotes the reproduced results of InfoGCN by [15]}
  \begin{tabular}{@{}l|cccc@{}}
    \toprule
   \multirow{3}*{Methods}  &\multicolumn{2}{c}{NTU RGB+D 60} &  \multicolumn{2}{c}{NTU RGB+D 120} \\
            & X-Sub(\%) & X-View(\%)   &  X-Sub(\%) & X-Set(\%)   \\
    \midrule
        MST-GCN  [2] &  89.0  & 95.1  & 82.8  & 84.5 \\
        InfoGCN [5] & 89.4* & 95.2*  & 84.2*  & 86.3*  \\
\hline
BlockGCN  
& 90.7 & 94.9 & 86.0 & 87.7\\
    \bottomrule
  \end{tabular}
  \label{tab:joint}
\end{table}

We further present the performance of our BlockGCN trained on each single modality. The experiment results for each modality on different benchmarks are provided in detail in Tab.~\ref{tab:b}. 

\begin{table}[h]
\centering
  \caption{Classification Accuracy of BlockGCN using Different Modalities on NTU RGB+D, NTU RGB+D 120, and Northwestern-UCLA Dataset.}
  \begin{tabular}{@{}l|ccccccc@{}}
    \toprule
       \multirow{2}*{Modality}  & \multicolumn{2}{c}{NTU-RGB+D 120} & \multicolumn{2}{c}{NTU-RGB+D} & \multirow{2}*{Northwestern-UCLA($\%$)}\\
       &  X-Sub(\%) & X-Set(\%)    & X-Sub(\%) & X-View(\%)   \\
    \midrule
    Joint & 86.0  & 87.7  & 90.7  & 94.9 & 92.5\\
    Bone & 87.3 &  88.7 & 90.9 &  95.1 & 93.3\\
    Motion &  82.4 &  84.3 & 88.5   & 92.9 & 91.2 \\
    Bone Motion & 82.6 &  84.4 &  88.6 & 92.7 & 90.7 \\
    \midrule
    Ensembled &89.7 & 90.9 & 92.8 & 96.4 & 96.8 \\
    \bottomrule
  \end{tabular}
\label{tab:b}
\end{table}

\subsection{Effect of normalization}
Applying L2 normalization on the Adjacency Matrices is widely adopted in GCN-based approaches. We found that L2 normalization benefits our Topological Invariance Encodings as well. As shown in \cref{tab:norm}, the smaller the dataset is, the more improvement L2 normalization brings. This shows that L2 normalization could alleviate the problem of overfitting. 

\begin{table}[h]
\centering
  \caption{The effect of normalization on NTU RGB+D, NTU RGB+D 120, and Northwestern-UCLA Dataset.}
  \begin{tabular}{@{}l|ccccccc@{}}
    \toprule
       \multirow{2}*{Modality} & L2 Norm & \multicolumn{2}{c}{NTU-RGB+D 120} & \multicolumn{2}{c}{NTU-RGB+D} & \multirow{2}*{UCLA($\%$)}\\
       &  X-Sub(\%) & X-Set(\%)    & X-Sub(\%) & X-View(\%)   \\
    \midrule
   \multirow{2}*{Ensembled} & -  &89.7 & 90.9 & 92.7 & 96.3 & 96.6 \\
    
    & \checkmark &89.7 & 90.9 & 92.8 & 96.4 & 96.8 \\
    \bottomrule
  \end{tabular}
\label{tab:norm}
\end{table}

\subsection{Effect of randomness}
To check the effect of randomness, we run our model on NTU-RGB+D 60\&120 using joint modality three times and report the results in Tab.~4. It can be seen that the standard deviations are relatively small and our model delivers stable performance.

\begin{table}[h]
    \centering
      \caption{The results of three different runs on NTU-RGB+D 60$\&$120 dataset using joint modality only.}
    \begin{tabular}{c|cccccc}
    \toprule
      Experiments & Modality  & 1 & 2 & 3 & mean & std \\
      \midrule
       NTU120 X-Sub & \multirow{4}*{Joint}& 86.0 & 85.6 & 86.0 & 85.87 & 0.19 \\
       NTU120 X-Set & & 87.7 & 88.0  & 88.1 &  87.93 & 0.17 \\
       NTU60 X-Sub & & 90.7 & 90.6 & 90.7 & 90.67 & 0.06 \\
       NTU60 X-View & & 94.9 &  94.8 & 94.5 & 94.73 & 0.17 \\
    \bottomrule
    \end{tabular}
    \label{tab:rand}
\end{table}

\subsection{Visualization of the learned weights}

The visualization of the learned Topological Invariance Encodings is shown in Fig.~1. It can be observed that these encodings are optimized to represent different levels of semantics at each layer according to the joint distances on the graph.

\begin{figure}[h]
     \centering
    \label{fig:topo}
    \begin{subfigure}[t]{0.19\textwidth}
         \centering
     \includegraphics[width=\textwidth]{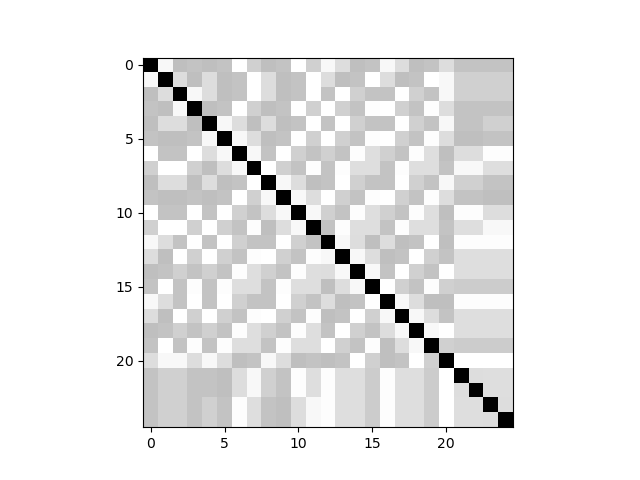}
     \caption{Layer 1.}
     \end{subfigure}
     \begin{subfigure}[t]{0.19\textwidth}
         \centering
         \includegraphics[width=\textwidth]{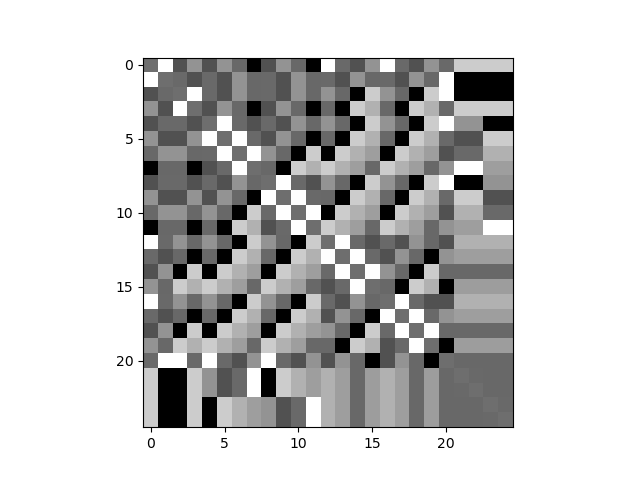}
         \caption{Layer 2.}
     \end{subfigure}
     \begin{subfigure}[t]{0.19\textwidth}
         \centering
         \includegraphics[width=\textwidth]{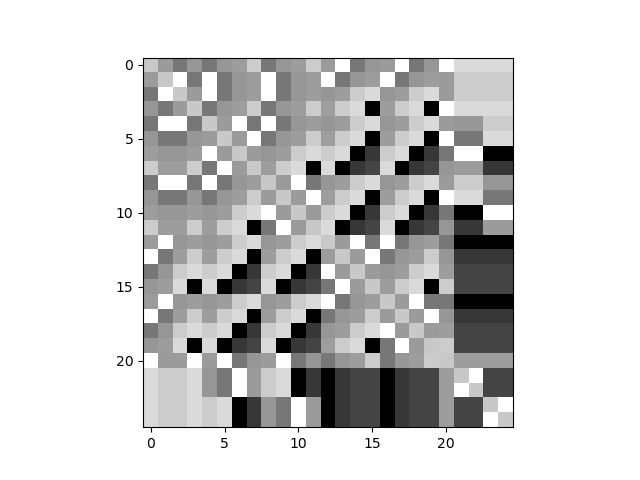}
         \caption{Layer 3.}
     \end{subfigure}
     \begin{subfigure}[t]{0.19\textwidth}
         \centering
         \includegraphics[width=\textwidth]{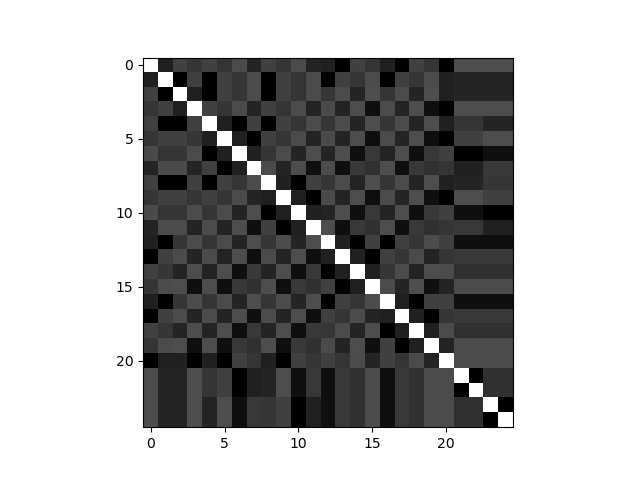}
         \caption{Layer 4.}
     \end{subfigure}
     \begin{subfigure}[t]{0.19\textwidth}
         \centering
         \includegraphics[width=\textwidth]{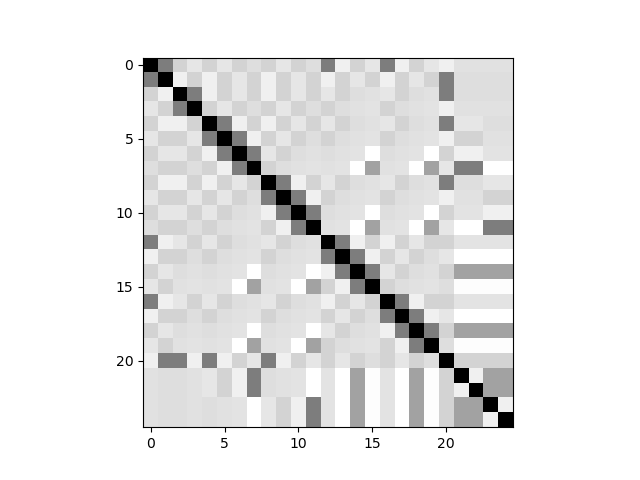}
         \caption{Layer 5.}
     \end{subfigure}
     \begin{subfigure}[t]{0.19\textwidth}
         \centering
         \includegraphics[width=\textwidth]{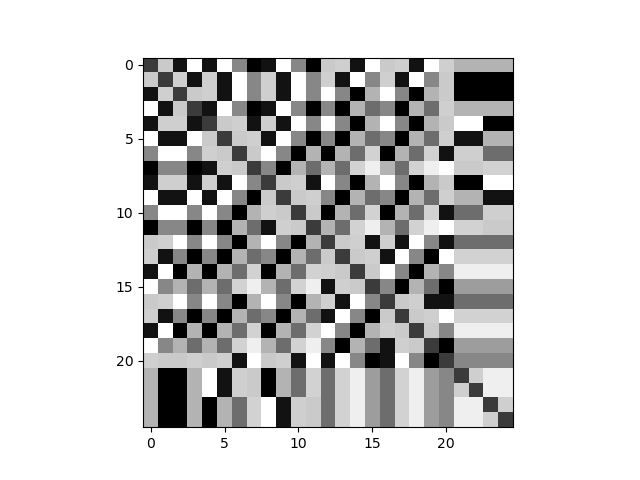}
         \caption{Layer 6.}
     \end{subfigure}
     \begin{subfigure}[t]{0.19\textwidth}
         \centering
         \includegraphics[width=\textwidth]{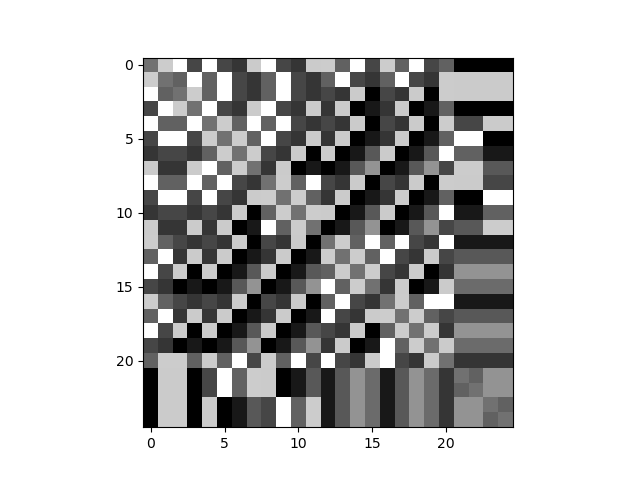}
         \caption{Layer 7.}
     \end{subfigure}
     \begin{subfigure}[t]{0.19\textwidth}
         \centering
         \includegraphics[width=\textwidth]{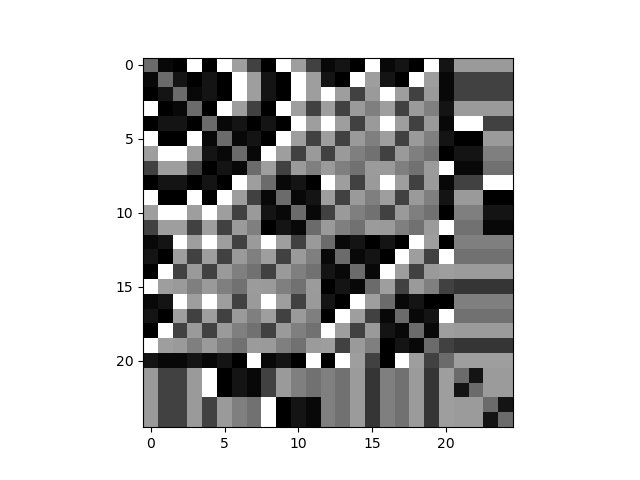}
         \caption{Layer 8.}
     \end{subfigure}
     \begin{subfigure}[t]{0.19\textwidth}
         \centering
         \includegraphics[width=\textwidth]{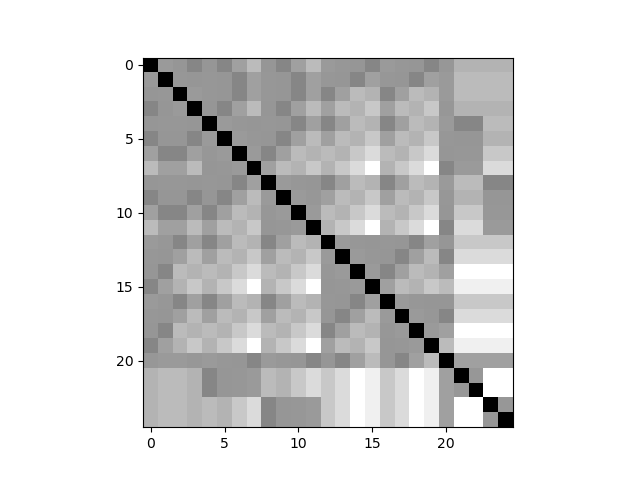}
         \caption{Layer 9.}
     \end{subfigure}
     \begin{subfigure}[t]{0.19\textwidth}
         \centering
         \includegraphics[width=\textwidth]{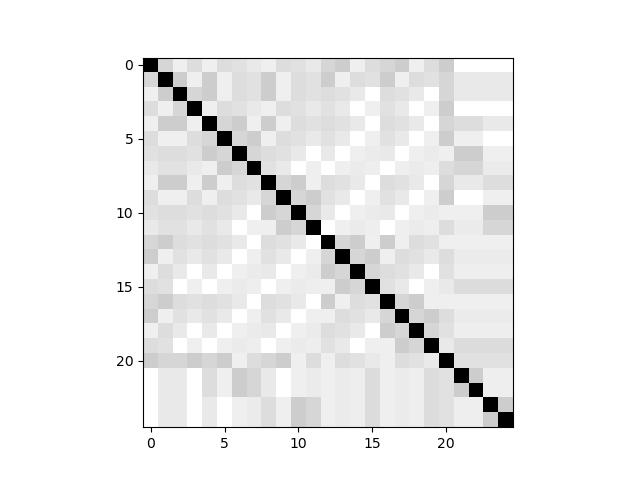}
         \caption{Layer 10.}
     \end{subfigure}
    \caption{The learned Topological Invariance Encodings of our BlockGCN at each layer. It can be seen that the learned weights are diverse and adapted to different levels of semantics.}
\end{figure}

To validate our analysis that the information of bone connectivity is lost after training. We also examined the learned weights of adjacency matrices at each layer of the GCN baseline model. The visualizations are provided in Fig.~2. As shown in the figure, the learned adjacency matrices are totally different from each other at each layer, although they are all initialized according to the bone connections.

\begin{figure}[t]
     \centering
    \label{fig:visual}
    \begin{subfigure}[t]{0.19\textwidth}
         \centering
     \includegraphics[width=\textwidth]{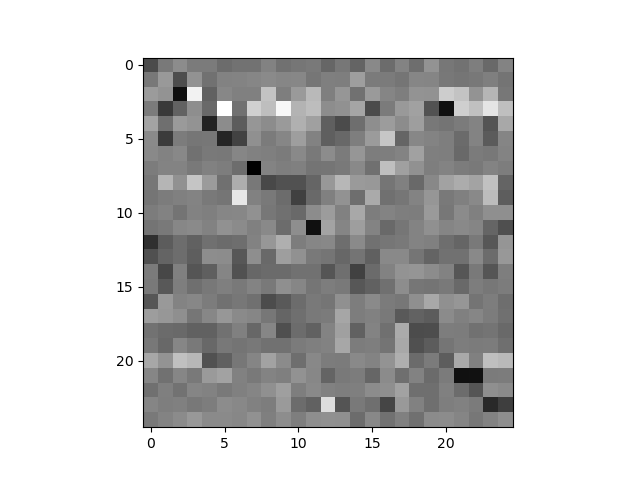}
     \caption{Layer 1.}
     \end{subfigure}
     \begin{subfigure}[t]{0.19\textwidth}
         \centering
         \includegraphics[width=\textwidth]{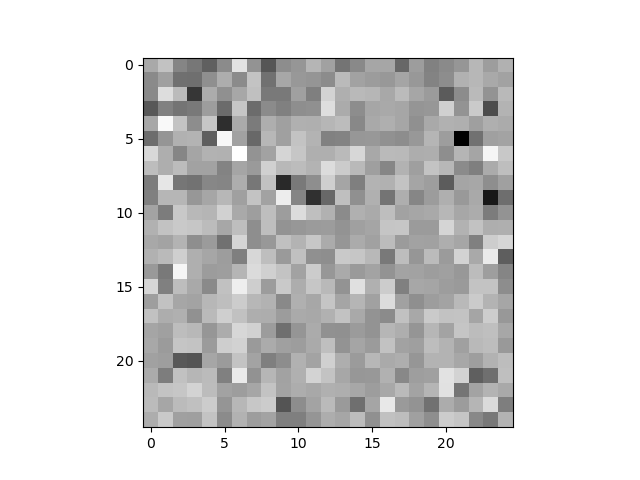}
         \caption{Layer 2.}
     \end{subfigure}
     \begin{subfigure}[t]{0.19\textwidth}
         \centering
         \includegraphics[width=\textwidth]{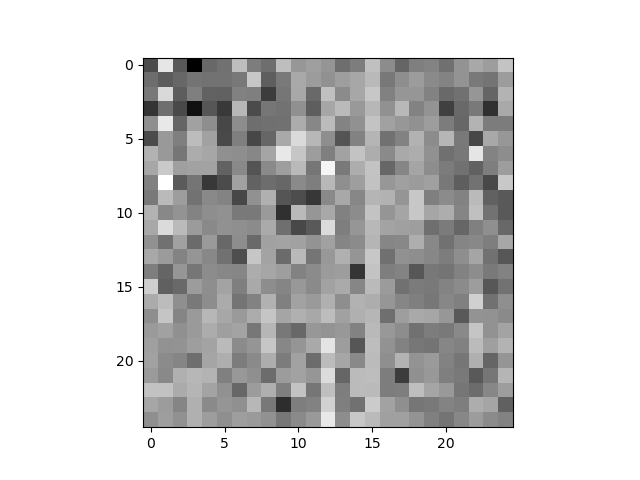}
         \caption{Layer 3.}
     \end{subfigure}
     \begin{subfigure}[t]{0.19\textwidth}
         \centering
         \includegraphics[width=\textwidth]{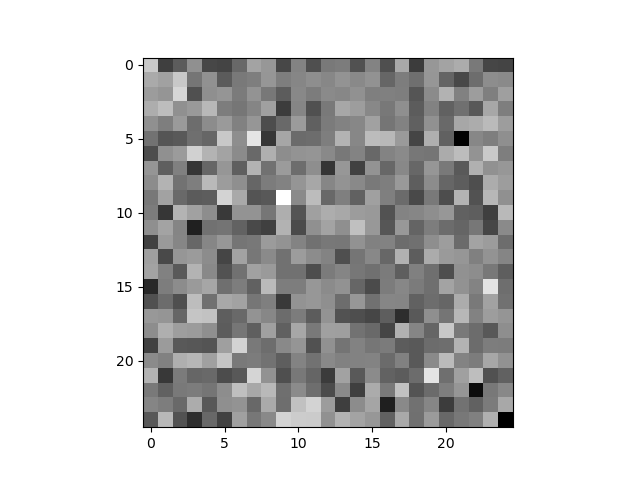}
         \caption{Layer 4.}
     \end{subfigure}
     \begin{subfigure}[t]{0.19\textwidth}
         \centering
         \includegraphics[width=\textwidth]{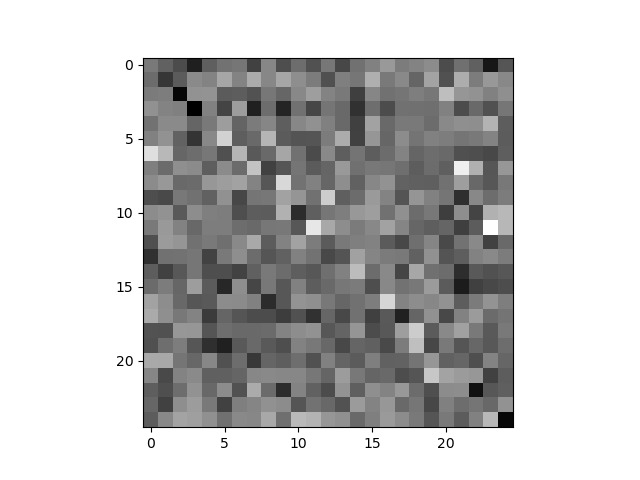}
         \caption{Layer 5.}
     \end{subfigure}
     \begin{subfigure}[t]{0.19\textwidth}
         \centering
         \includegraphics[width=\textwidth]{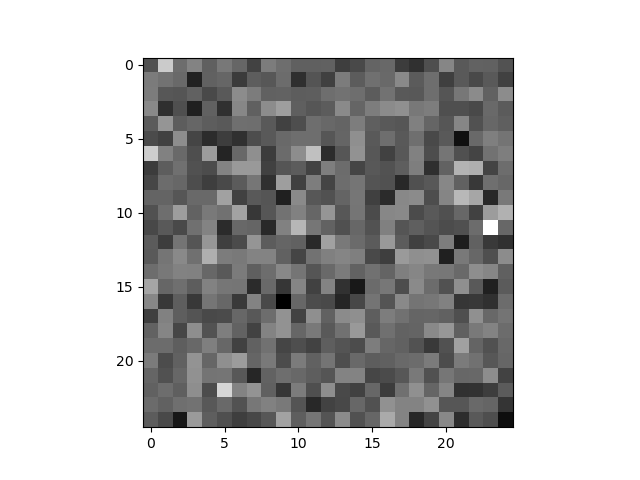}
         \caption{Layer 6.}
     \end{subfigure}
     \begin{subfigure}[t]{0.19\textwidth}
         \centering
         \includegraphics[width=\textwidth]{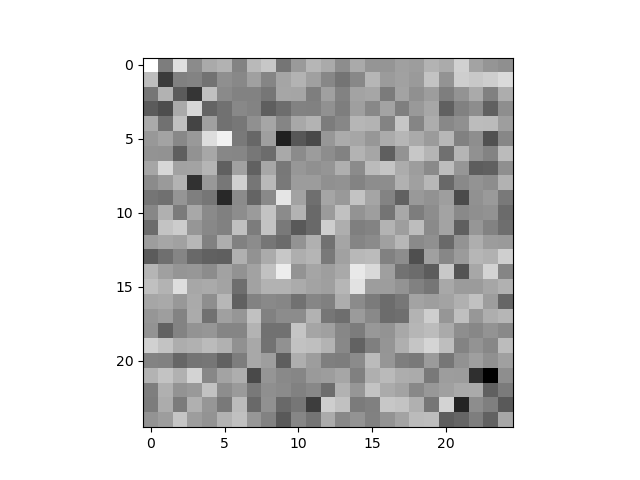}
         \caption{Layer 7.}
     \end{subfigure}
     \begin{subfigure}[t]{0.19\textwidth}
         \centering
         \includegraphics[width=\textwidth]{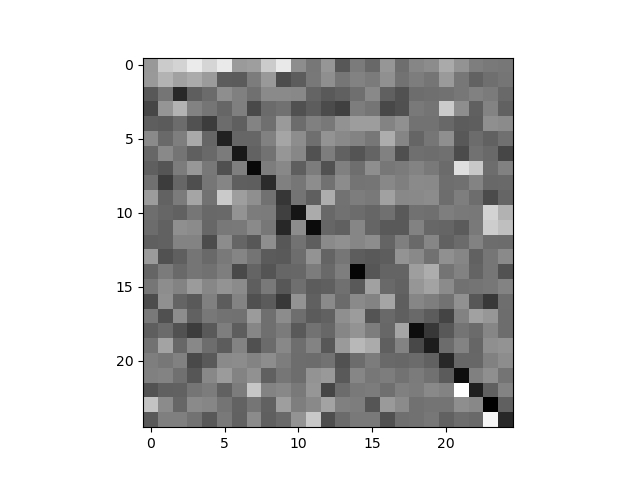}
         \caption{Layer 8.}
     \end{subfigure}
     \begin{subfigure}[t]{0.19\textwidth}
         \centering
         \includegraphics[width=\textwidth]{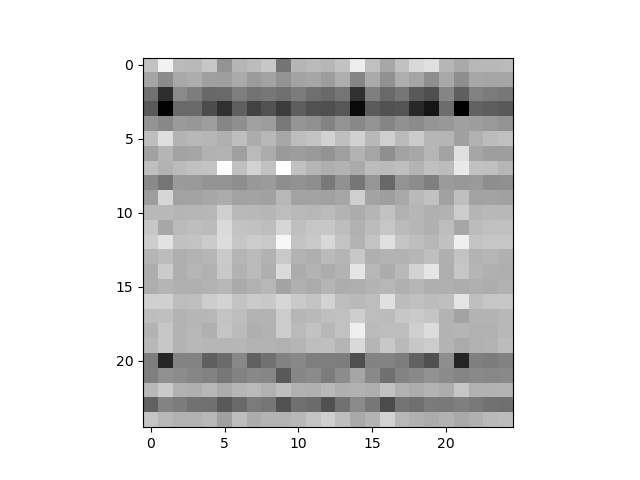}
         \caption{Layer 9.}
     \end{subfigure}
     \begin{subfigure}[t]{0.19\textwidth}
         \centering
         \includegraphics[width=\textwidth]{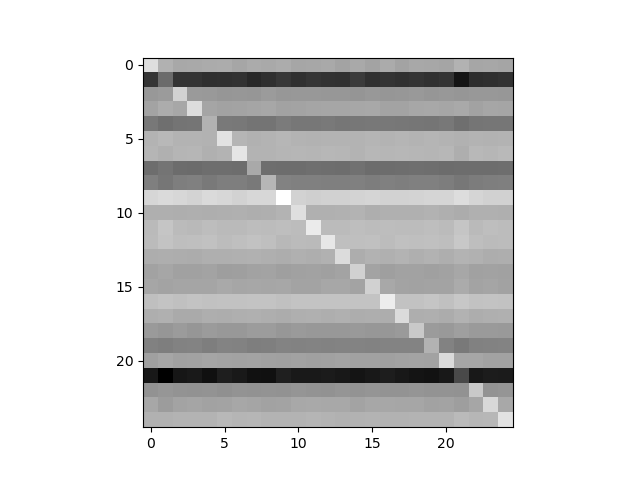}
         \caption{Layer 10.}
     \end{subfigure}
        \begin{subfigure}[t]{0.3\textwidth}
         \centering
         \includegraphics[width=\textwidth]{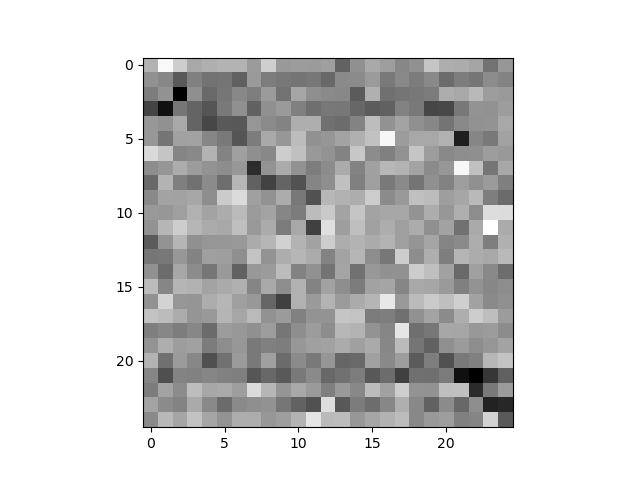}
         \caption{Mean of the learned A.}
     \end{subfigure}
     \begin{subfigure}[t]{0.3\textwidth}
         \centering
         \includegraphics[width=\textwidth]{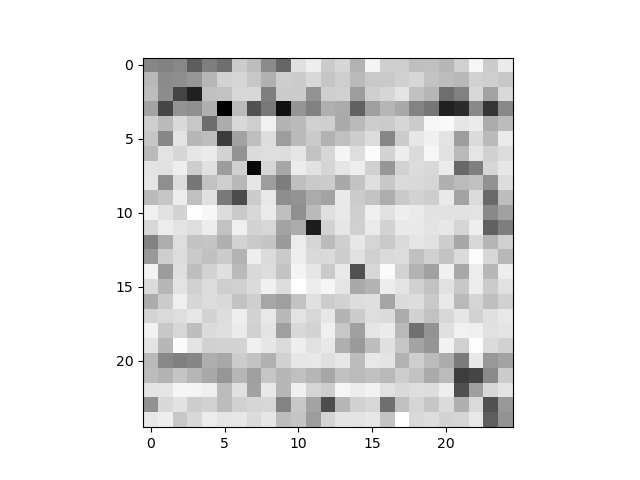}
         \caption{Standard Deviation of the learned A.}
     \end{subfigure}
     \begin{subfigure}[t]{0.3\textwidth}
         \centering
         \includegraphics[width=\textwidth]{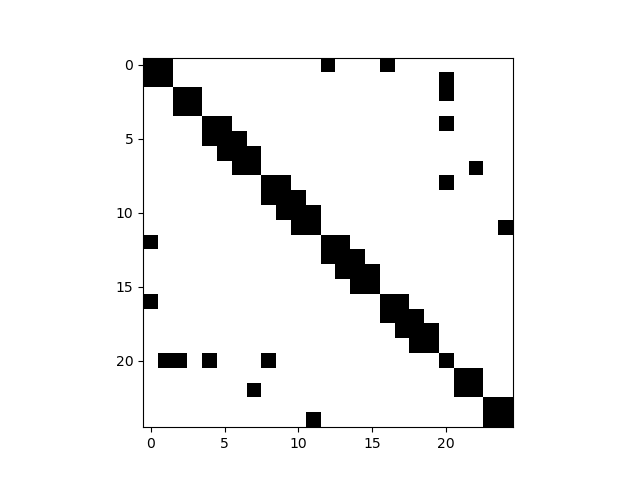}
         \caption{Bone connections.}
     \end{subfigure}
    \caption{The learned adjacency matrices of the GCN baseline model at each layer (Darker colors stand for larger weights). It can be seen that the learned weights vary dramatically among different layers and deviate far from the bone connections, which are used for initialization.}
\end{figure}

\section{Hyperparameters}
We provide the default hyperparameters used for training our BlockGCN on the NTU RGB+D, NTU RGB+D 120, and Northwestern-UCLA datasets. Throughout our paper, we consistently train a 10-layer model with a maximum of 256 channel dimensions. \cref{tab:a} presents the default hyperparameters for our BlockGCN on these datasets:

\begin{table}[t]
  \caption{Default Hyperparameters for BlockGCN on NTU RGB+D, NTU RGB+D 120, and Northwestern-UCLA.}
    \centering
    %\resizebox{.95\columnwidth}{!}{
    \begin{tabular}{l|cc}
       \toprule
        Config. & NTU RGB+D and NTU RGB+D 120 & Northwestern-UCLA \\
  \midrule
      random choose & False & True \\
      random rotation & True & False \\
      window size & 64 & 52\\
      weight decay & 4e-4 & 2e-4 \\
      base lr & 0.1 & 0.1 \\
      lr decay rate & 0.1 & 0.1 \\
      lr decay epoch & 110, 120 & 90 100 \\
      warm up epoch & 5 & 5 \\
      batch size & 64 & 16 \\
      num. epochs & 140 & 120\\
      optimizer & Nesterov Accelerated Gradient & Nesterov Accelerated Gradient \\
       \hline
    \end{tabular}
    \label{tab:a}
\end{table}

\end{document}